\documentclass[11pt]{article}
\usepackage{acl}
\usepackage{times}
\usepackage{latexsym}
\usepackage[T1]{fontenc}
\usepackage[utf8]{inputenc}
\usepackage{microtype}
\usepackage{inconsolata}
\usepackage{graphicx}
\usepackage{booktabs}
\usepackage{multirow}
\usepackage{amsmath}
\usepackage{amssymb}
\usepackage{xcolor}
\usepackage{subcaption}
\usepackage{xcolor}
\newcommand{\ci}[1]{{\scriptsize\textcolor{gray}{[#1]}}}
\newcommand{\up}[1]{\textcolor{green!60!black}{\textbf{+#1}}}
\newcommand{\down}[1]{\textcolor{red}{\textbf{--#1}}}

\newcommand{\wone}{W_1}
\newcommand{\dr}{\sigma_r}

\title{The Limits of Simulated Societies: How Post-Training and Survey Fine-Tuning Erase Cross-Cultural Variance}

\author{Rojin Ziaei \\
  Georgetown University \\
  \texttt{nz204@georgetown.edu}\\}

\begin{document}
\maketitle

\begin{abstract}

Using large language models (LLMs) to simulate diverse human populations has the potential to transform many aspects of computational social science, yet many evaluations score the average response rather than the spread of opinion within real groups. Here, we develop a diagnostic framework that measures point accuracy alongside dispersion retention, the ratio of predicted to human standard deviation ($\dr$), on 10{,}000 respondent--question pairs from the World Values Survey (WVS) spanning twelve countries and six continents. We evaluate eleven zero-shot language models and five variants fine-tuned on WVS data with SFT, DPO, and GRPO. We identify a failure mode we term \textit{consensus collapse}, where alignment training compresses outputs toward one stereotype per group. Along the post-training trajectory from the Llama~3.1 70B base to the Tulu~3 checkpoints, the first stage, supervised instruction tuning, removes half of the spread with minimal accuracy gain ($\dr$ 1.22 to 0.59; accuracy $+0.9$ points), the later stages do not restore it, and a gap opens between WEIRD and non-WEIRD countries that survey fine-tuning then deepens while pursuing higher point accuracy. The most accurate model (Tulu~3 70B-DPO fine-tuned on WVS, 57.9\%) keeps half the human spread overall ($\dr = 0.50$) and 11\% of it for Nigeria, against 0.70--0.87 for WEIRD countries. Raising the sampling temperature to 1.0 leaves the Wasserstein-1 distance ($\wone$) to human distributions unchanged for both fine-tuned DPO models, and GRPO on Qwen~3.5 9B does not restore the spread under either an accuracy reward or a distribution-shaped reward. Mixing the aligned model with an unaligned prior raises $\dr$ from 0.51 to 0.62 on a held-out split but leaves Nigeria at 0.36. Point accuracy alone therefore misjudges these simulators, and current post-training trades diversity for consensus.

\end{abstract}


\section{Introduction}
\label{sec:intro}

\begin{figure}[t]
\centering
\fbox{\includegraphics[width=0.97\columnwidth]{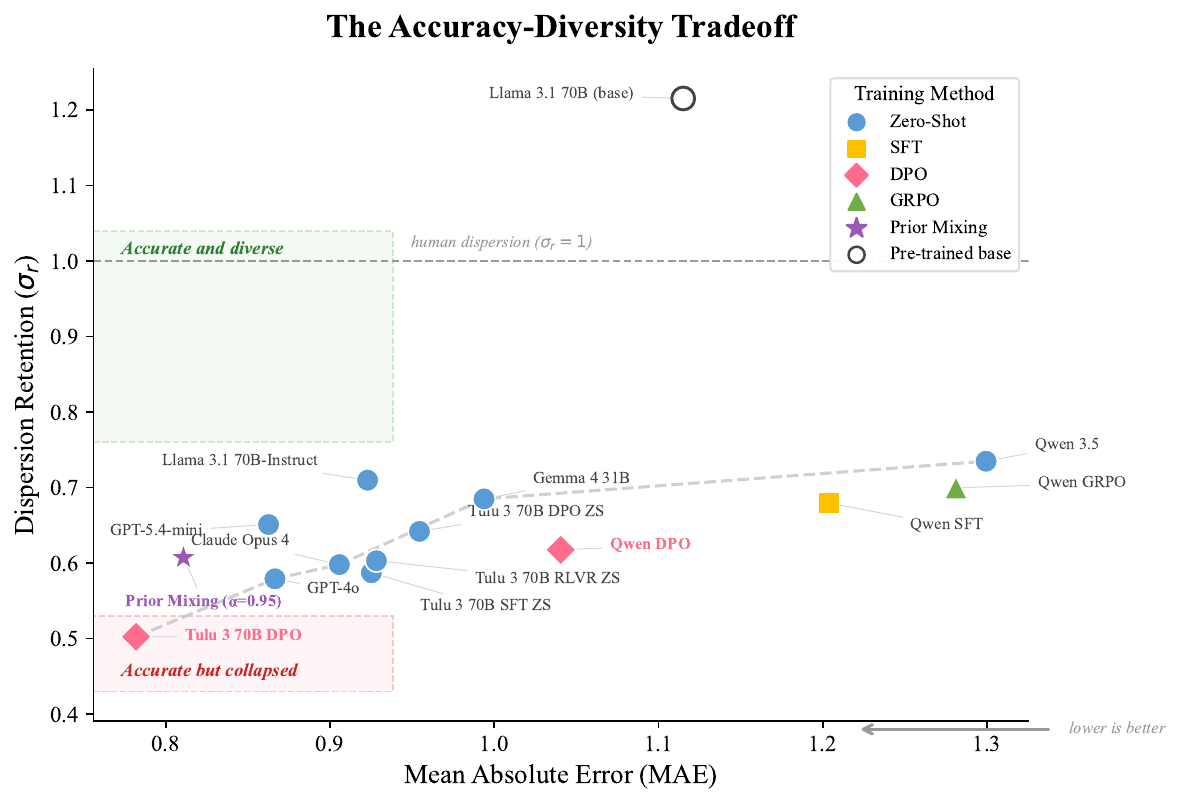}}
\caption{\textbf{The Accuracy--Diversity Tradeoff.} Each point is a model evaluated on 10{,}000 held-out WVS responses ($x$: mean absolute error; $y$: dispersion retention $\dr$). The pre-trained Llama~3.1 70B base (hollow circle) is the only model above the human-dispersion line ($\dr = 1$); every post-trained checkpoint falls below it, and alignment fine-tuning (pink diamonds) pushes models into the ``accurate but collapsed'' region---Tulu~3 70B-DPO (+WVS) achieves the lowest MAE but keeps only half of the human spread. Prior mixing (purple star; the dev-selected $\alpha = 0.95$ applied to the full evaluation set) recovers part of the spread at a small accuracy cost. No model reaches the ``accurate and diverse'' quadrant.}
\label{fig:tradeoff}
\end{figure}

The application of large language models (LLMs) as computational stand-ins for human survey respondents represents a major frontier in computational social science. By accurately simulating the responses of diverse demographic groups, LLMs enable subpopulation representative modeling (SRM) \citep{simmons2024srm}, a framework that allows researchers to pilot interventions, stress-test public policies, and generate synthetic panels without the financial and temporal bottleneck of traditional cohort recruitment \citep{horton2023large, manning2024automated}. This potential is increasingly supported by empirical advances. Baseline zero-shot models utilizing demographic personas have already demonstrated the ability to reproduce aggregate public opinion \citep{argyle2023out, rss2024silicon}. Fine-tuning on individual-level responses improves this fidelity further: \citet{kolluri2025simulacra} fine-tune on 2.9 million responses from 210 social science experiments and report that their strongest model's predictions on unseen studies are 26\% more aligned with human response distributions than its base model's.  


As promising as this may seem, a growing body of evidence warns that the paradigm carries fundamental risks. \citet{wang2024flatten} demonstrate that LLMs prompted with demographic identities harmfully \textit{flatten and misportray} those identity groups, collapsing rich within-group diversity into monolithic stereotypes. \citet{krsteski2025valid} find that pure LLM synthesis introduces substantial bias (24--86\%), necessitating post-hoc ``rectification'' using held-out human data. \citet{qu2024bias} confirm this on the World Values Survey specifically, showing that performance degrades sharply for non-Western, non-English-speaking populations. This contrast, with the promise of alignment on one hand and the evidence of systematic flattening on the other, remains unreconciled because it has been difficult to measure the precise mechanism by which alignment objectives distort the distributional properties of model outputs.

\citet{park2024generative} show that these failures can be overcome when models perform modeling grounded in two-hour qualitative interviews: their agents replicate participants' General Social Survey answers 85\% as accurately as the participants replicate their own answers two weeks later. However, such data is prohibitively expensive for the vast majority of global survey datasets, and most cross-national survey simulation research operates with structured demographic personas rather than biographical transcripts. Prior work establishes that LLMs flatten demographic groups under persona prompting. What is not yet established is whether existing LLM training paradigms mitigate or amplify this flattening. 

Using data from the World Values Survey (WVS) across 12 culturally diverse countries, we construct value-encoded personas that explicitly encode the dimensions of the Inglehart-Welzel cultural-map \citep{inglehart2006mapping}. We then use these personas to evaluate eleven off-the-shelf checkpoints zero-shot and five variants fine-tuned on WVS data with three methods (SFT, DPO, GRPO). To execute this evaluation, we develop a distributional diagnostic framework built on four complementary metrics, accuracy, MAE, Wasserstein-1 distance ($\wone$), and deviation ratio ($\dr$), that captures both point-prediction quality and dispersion retention. This framework enables us to characterize a systematic failure mode we call \textit{consensus collapse}: the structural compression of model output variance induced by alignment fine-tuning.

We make three contributions and propose one practical mitigation. First, we formally define and measure consensus collapse, showing that general-purpose post-training already amplifies the flattening observed by \citet{wang2024flatten} and that our survey fine-tuning deepens it further, where the best-performing fine-tuned model keeps only half of the human response spread ($\dr = 0.50$). Second, we show that the collapse is not recoverable by raising the sampling temperature: for two Direct Preference Optimization (DPO) \citep{rafailov2023direct} models of different families and scales, sampling at up to $T = 1.0$ leaves $\wone$ unchanged. Third, we reveal that the collapse is inequitable: Nigeria keeps only 11\% of the human response spread ($\dr = 0.11$) compared to 70--87\% for WEIRD countries, and that Group Relative Policy Optimization (GRPO) \citep{shao2024deepseekmath} on Qwen~3.5 9B fails to restore it under either an accuracy reward or a distribution-shaped reward. As a partial mitigation, we introduce prior mixing, a post-hoc probe that probabilistically mixes an aligned model with an unaligned prior, raising $\dr$ from 0.51 to 0.62 and reducing $\wone$ from 0.54 to 0.51 on a held-out split while underscoring that the underlying collapse resists simple correction.

\section{Related Work}
\label{sec:related}
\paragraph{Subpopulation Representative Modeling.} The idea of using LLMs as ``Homo Silicus'' computational stand-ins for human economic and social agents was formalized by \citet{horton2023large}. \citet{manning2024automated} extend this to automated hypothesis generation and testing. \citet{simmons2024srm} formalize SRM, evaluating the ability of LLMs to generalize from empirical data via in-context learning. \citet{kim2023augmented} fine-tune on the General Social Survey and report an AUC of 0.86 for retrodicting individual opinions in unsurveyed years, with weaker performance on entirely unasked questions.

\paragraph{Value-encoded persona Prompting and Flattening.} A common approach to SRM involves conditioning LLMs on demographic descriptors. The ``Random Silicon Sampling'' framework \citep{rss2024silicon} demonstrates that group-level demographic prompting alone can reflect U.S.\ public opinion distributions. However, both \citet{wang2024flatten} and \citet{qu2024bias} reveal failure modes, where LLMs systematically flatten within-group diversity into stereotypes, and performance degrades sharply for non-Western populations. We extend this line of work by quantifying the WEIRD/non-WEIRD asymmetry directly: our per-country diagnostics (\S\ref{sec:country}) show that dispersion retention for WEIRD countries ($\sigma_r = 0.70$--$0.87$) is up to $8\times$ higher than for non-WEIRD countries ($\sigma_r = 0.11$--$0.60$), and that at 70B alignment fine-tuning widens rather than narrows this gap.

\paragraph{Fine-Tuning for Survey Simulation.} \citet{kolluri2025simulacra} construct SOCSCI210, a dataset of 2.9 million responses from 210 experiments, and show that Supervised Fine-Tuning (SFT) \citep{ouyang2022training} meaningfully improves distributional alignment. Their Demographic Contrastive DPO approach, which pairs respondents with contrasting demographics who answered the same question differently, is the basis for our DPO pipeline. We extend their work by systematically diagnosing why alignment collapses variance, by evaluating on a cross-national rather than U.S.-only benchmark, and by testing whether an alternative training objective offers an escape.

\paragraph{Post-Hoc Calibration and Rectification.} \citet{krsteski2025valid} demonstrates that pure LLM synthesis introduces 24--86\% bias in population estimates. They propose \textit{rectification} which is importance weighting using held-out human responses. Our prior-mixing probe (\S\ref{sec:mitigation}) addresses the same problem without requiring ground-truth human data at inference time.

\paragraph{Alignment Bias.} \citet{santurkar2023whose} show that human-feedback-tuned LLMs lean left relative to US demographic groups and poorly reflect the views of some groups, such as respondents aged 65 and over. \citet{sharma2024towards} show that the human preference judgments used in RLHF favor sycophantic responses. \citet{durmus2024measuring} show that LLM responses to cross-national survey questions are more similar to the opinions of populations such as those of the USA and some European and South American countries than to others. Our work moves from documenting this skew to characterizing how it scales with alignment. Comparing unaligned and aligned variants of the same model family, we find that at 70B alignment fine-tuning \textit{amplifies} the WEIRD/non-WEIRD gap rather than mitigating it ($\sigma_r = 0.70$--$0.87$ for WEIRD vs.\ $0.11$--$0.60$ for non-WEIRD countries under DPO), whereas the 9B Qwen family shows only a small gap at every training stage (\S\ref{sec:country}).

\paragraph{Epistemic Modeling.} \citet{park2024generative} proposes using agents built from two-hour interview transcripts, demonstrating this method can match participants' own two-week test--retest consistency 85\% of the way. This suggests a spectrum of input richness, including demographics through structured personas to dense biographical context, along which consensus collapse may progressively dissolve. Our work provides the diagnostic evidence at the structured persona point on this spectrum, where most cross-national survey simulation research currently operates.

\section{Experimental Setup}
\label{sec:setup}

\subsection{Dataset}
\label{sec:dataset}

Each data record, drawn from WVS Wave~7 \citep{haerpfer2022world}, pairs a respondent's \textit{value-encoded persona} (a natural-language description of 59--70 words, synthesized from
each respondent's demographic variables and derived cultural indices in the WVS microdata) with a question and the respondent's ground-truth ordinal answer. Notably, each persona encodes not only standard demographics (age, gender, country) but also the two Inglehart--Welzel cultural-map axes (traditional vs.\ secular-rational values and survival vs.\ self-expression orientation), alongside interpersonal trust, civic engagement, and religious identity.

\begin{figure}[t]
\centering
\small
\fbox{\parbox{0.93\columnwidth}{\textit{I am a 68-year-old male from Australia. My worldview is strongly traditional and authority-respecting, and I am transitioning from survival priorities toward self-expression values. In my interactions with others, I am generally trusting of others and open to strangers. Socially, I am moderately active in community life with selective group memberships. My identity is deeply tied to being Roman Catholic.}}}
\caption{A representative value-encoded persona (verbatim from the dataset, $\sim$60 words). It encodes the two Inglehart--Welzel dimensions (tradition/secular, survival/self-expression), interpersonal trust, civic embeddedness, and religious identity, drawing on the core constructs that the WVS instrument captures.}
\label{fig:persona}
\end{figure}

\subsection{Question Split Strategy}
\label{sec:question_split}

To test whether models can generalize persona-grounded reasoning to unseen value domains, we partition the 21 survey questions into two non-overlapping sets:

\begin{table}[t]
\centering
\small
\begin{tabular}{llp{3.5cm}}
\toprule
\textbf{Set} & \textbf{Count} & \textbf{Purpose} \\
\midrule
Train & 15 & Fine-tuning signal \\
Test  & 6  & Held-out evaluation \\
\bottomrule
\end{tabular}
\caption{Question split. Training questions: Q2, Q6, Q27, Q46--Q50, Q58--Q60, Q106, Q164, Q185, Q240. Test questions: Q1, Q29, Q64, Q71, Q182, Q189. The test set spans the Values, Social Trust, and Political clusters; the Economic cluster appears only in training.}
\label{tab:split}
\end{table}

The test set deliberately includes high-consensus questions (Q1), contested questions (Q182), and culturally sensitive questions (Q189). We use 100{,}000 respondent--question pairs for training (50{,}000 for GRPO), drawn with a fixed seed from all 12~countries, and 10{,}000 for testing.

\subsection{Prompt Format}
All chat models receive an identical three-turn chat prompt instructing them to simulate a survey respondent given a persona and to return a single integer on the question's response scale; the pre-trained base model, which has no chat template, receives the same content as a plain completion prompt. Both templates are in Appendix~\ref{app:prompt}.

We evaluate eleven zero-shot language models (four proprietary API models, open-weight instruction-tuned models from 9B to 70B, and the four checkpoints of the Llama~3.1 70B $\to$ Tulu~3 post-training trajectory \citep{lambert2025tulu3pushingfrontiers}: the pre-trained base, Tulu~3 70B-SFT, Tulu~3 70B-DPO, and Tulu~3 70B-RLVR, the final checkpoint trained with reinforcement learning with verifiable rewards) and five WVS-fine-tuned variants (Qwen 3.5 9B with SFT, with DPO, and with GRPO under two reward designs, and Tulu 3 70B with DPO; \S\ref{sec:grpo}); full configurations including LoRA rank, learning rates, and precision are provided in Appendix~\ref{app:models}.

\subsection{Evaluation Metrics}
\label{sec:metrics}
We employ four complementary metrics. Accuracy is the exact-match rate between prediction and ground-truth ordinal answer, and Mean Absolute Error (MAE) is the average magnitude of prediction errors on the ordinal scale; both are pooled over all test pairs. The two distributional metrics are computed per (country, question) cell and averaged over cells. We use Wasserstein-1 Distance ($\wone$) to measure the distance between ground-truth and predicted population distributions ($\downarrow$ better). Deviation Ratio ($\dr$) is $\sigma_{\text{pred}} / \sigma_{\text{truth}}$, the ratio of predicted to human standard deviation, which we also call dispersion retention; $\dr = 1.0$ is perfect, $\dr \to 0$ is mode collapse, and $\dr > 1$ is over-dispersion, so values are read by their distance from 1. Because it is a ratio of standard deviations, $\dr = 0.50$ keeps half the spread and a quarter of the variance.
\subsection{Inference}
\label{sec:inference}

All primary evaluations use greedy decoding ($T = 0$) to minimize variation in the outputs across the 10,000 test pairs. Because greedy decoding could in principle exaggerate variance compression by collapsing each prediction to the argmax token, we additionally evaluate both WVS-fine-tuned DPO models, Tulu 3 70B-DPO ($\sigma_r = 0.50$) and Qwen 3.5 9B-DPO ($\sigma_r = 0.62$), at $T = 0.5, 0.7,$ and $1.0$, spanning the range from sharpened sampling to the model's untransformed output distribution. This sweep tests whether consensus collapse is a decoding artifact that broader sampling would dissolve (\S\ref{sec:temp}).

\section{Results}
\label{sec:results}

\subsection{Alignment reduces human variance}
\label{sec:main_results}

All metrics include 95\% bootstrap confidence intervals (1{,}000 iterations, resampling over $Q \times C$ cells). Table~\ref{tab:main} presents aggregate results across the full test set of 10{,}000 predictions.


\begin{table*}[t]
\centering
\small
\resizebox{\textwidth}{!}{%
\begin{tabular}{lcccc cc}
\toprule
\textbf{Model} & \textbf{Acc\%} & \textbf{MAE} & $\mathbf{\wone}$ $\downarrow$ & $\mathbf{\dr}$ ($\rightarrow 1$)
& $\Delta$\textbf{Acc} & $\Delta\dr$ \\
\midrule
\multicolumn{7}{l}{\textit{Zero-Shot Baselines}} \\
GPT-4o & 52.8 \ci{46.2, 59.6} & 0.87 \ci{0.68, 1.08} & 0.59 \ci{0.48, 0.73} & 0.58 \ci{0.45, 0.71} & --- & --- \\
GPT-5.4-mini & 51.0 \ci{44.6, 57.6} & 0.86 \ci{0.69, 1.04} & 0.57 \ci{0.46, 0.71} & 0.65 \ci{0.54, 0.76} & --- & --- \\
Tulu 3 70B-SFT & 51.2 \ci{44.5, 58.4} & 0.93 \ci{0.73, 1.13} & 0.71 \ci{0.55, 0.87} & 0.59 \ci{0.46, 0.71} & --- & --- \\
Llama 3.1 70B-Instruct & 50.7 \ci{44.5, 57.4} & 0.92 \ci{0.74, 1.13} & 0.60 \ci{0.48, 0.75} & 0.71 \ci{0.58, 0.85} & --- & --- \\
Tulu 3 70B-RLVR & 50.6 \ci{43.9, 57.6} & 0.93 \ci{0.74, 1.13} & 0.69 \ci{0.54, 0.85} & 0.60 \ci{0.48, 0.73} & --- & --- \\
Llama 3.1 70B (base) & 50.3 \ci{45.1, 56.1} & 1.12 \ci{0.92, 1.32} & 0.52 \ci{0.42, 0.66} & 1.22 \ci{1.08, 1.43} & --- & --- \\
Tulu 3 70B-DPO & 49.7 \ci{43.1, 56.8} & 0.95 \ci{0.75, 1.16} & 0.69 \ci{0.55, 0.86} & 0.64 \ci{0.52, 0.78} & --- & --- \\
Gemma 4 31B & 49.8 \ci{43.1, 56.4} & 0.99 \ci{0.78, 1.23} & 0.72 \ci{0.55, 0.90} & 0.69 \ci{0.58, 0.81} & --- & --- \\
Claude Opus 4 & 48.2 \ci{41.5, 55.0} & 0.91 \ci{0.73, 1.10} & 0.68 \ci{0.54, 0.83} & 0.60 \ci{0.49, 0.72} & --- & --- \\
Gemini 2.5 Pro & 43.0 \ci{37.0, 49.1} & 1.28 \ci{0.97, 1.63} & 0.89 \ci{0.64, 1.20} & 0.87 \ci{0.72, 0.99} & --- & --- \\
Qwen 3.5 9B & 38.4 \ci{31.5, 45.8} & 1.30 \ci{1.00, 1.67} & 1.01 \ci{0.77, 1.32} & 0.73 \ci{0.62, 0.93} & --- & --- \\
\midrule
\multicolumn{7}{l}{\textit{Fine-Tuned on WVS}} \\
\textbf{Tulu 3 70B-DPO (+WVS)} & \textbf{57.9} \ci{52.0, 63.8} & \textbf{0.78} \ci{0.62, 0.96} & 0.55 \ci{0.44, 0.68} & 0.50 \ci{0.39, 0.62} & \up{8.2} \ci{7.5, 8.8} & \down{0.14} \ci{-0.17, -0.13} \\
Qwen 3.5 9B-SFT (+WVS) & 52.9 \ci{46.5, 59.6} & 1.20 \ci{0.90, 1.59} & 0.89 \ci{0.64, 1.22} & 0.68 \ci{0.54, 0.86} & \up{14.5} \ci{13.5, 15.4} & \down{0.05} \ci{-0.16, -0.02} \\
Qwen 3.5 9B-DPO (+WVS) & 53.0 \ci{47.2, 59.0} & 1.04 \ci{0.78, 1.32} & 0.73 \ci{0.54, 0.96} & 0.62 \ci{0.50, 0.78} & \up{14.6} \ci{13.7, 15.4} & \down{0.11} \ci{-0.21, -0.08} \\
Qwen 3.5 9B-GRPO (+WVS) & 38.6 \ci{31.7, 45.9} & 1.28 \ci{0.99, 1.64} & 1.00 \ci{0.76, 1.31} & 0.70 \ci{0.59, 0.90} & \up{0.2} \ci{-0.2, 0.5} & \down{0.03} \ci{-0.05, -0.02} \\
Qwen 3.5 9B-GRPO-dist (+WVS) & 47.5 \ci{40.3, 54.5} & 1.01 \ci{0.78, 1.28} & 0.84 \ci{0.67, 1.07} & 0.21 \ci{0.14, 0.27} & \up{9.1} \ci{8.1, 10.1} & \down{0.52} \ci{-0.64, -0.50} \\
\bottomrule
\end{tabular}}
\caption{\textbf{Diagnostic evaluation across 10{,}000 held-out predictions (6~questions $\times$ 12~countries).} Bracketed values report 95\% cluster-bootstrap confidence intervals (1{,}000 iterations over the 71 country--question cells). \textbf{WVS Fine-Tuning Shift} ($\Delta\text{Acc}$, $\Delta\dr$) measures the change relative to each model's corresponding zero-shot checkpoint (Tulu~3 70B-DPO for the 70B run; Qwen~3.5 9B for the 9B runs), with 95\% paired-bootstrap CIs. The best accuracy and MAE are bolded. While WVS fine-tuning via SFT and DPO substantially improves point accuracy ($+8.2$ to $+14.6$ points), it systematically compresses dispersion ($\Delta\dr$ of $-0.05$ to $-0.14$). Sparse-reward GRPO leaves accuracy unchanged and $\dr$ within 0.03 of the baseline, whereas distribution-shaped GRPO improves accuracy ($+9.1$) at the cost of severe variance collapse ($\dr = 0.21$). Unparseable responses are excluded (Gemini~2.5 Pro: $n = 8{,}377$; Llama~3.1 70B Base: $n = 9{,}988$; all other models: $n = 10{,}000$).}
\label{tab:main}
\end{table*}

The results reveal a trade-off. The most accurate model (Tulu~3 70B-DPO fine-tuned on WVS, 57.9\%) keeps only half of the human response spread ($\dr = 0.50$), whereas the unaligned Qwen~3.5 9B keeps 73\% of it but reaches only 38.4\% accuracy. Two reference points calibrate these accuracies: always predicting each question's modal answer scores 52.6\%, level with GPT-4o, and predicting the modal answer of each (country, question) cell scores 63.2\% in-sample, with $\dr = 0$ by construction. Every model sits between these two, so fine-tuning moves models toward the majority-class predictor without reaching it, shedding spread along the way. DPO compresses the spread below the unaligned baseline at both scales (paired $\Delta\dr$ of $-0.11$ and $-0.14$) and SFT at 9B compresses it slightly ($-0.05$), every CI excluding zero (Table~\ref{tab:main}); sparse-reward GRPO leaves accuracy unchanged and $\dr$ within 0.03 of the baseline (\S\ref{sec:grpo}). Fine-tuning does improve location: MAE and $\wone$ fall for every SFT and DPO run (Tulu $\wone$ 0.69 $\to$ 0.55; Qwen 1.01 $\to$ 0.89 with SFT and 0.73 with DPO), so consensus collapse names the loss of spread, not a loss of distributional accuracy overall. The pre-trained Llama~3.1 70B base, prompted in completion style, is the only over-dispersed model ($\dr = 1.22$ at 50.3\% accuracy) and has the lowest $\wone$ in the table (0.52); every post-trained checkpoint, whether post-trained by its provider or fine-tuned on WVS in this work, falls below 1. Along the Tulu trajectory (Table~\ref{tab:trajectory}), general post-training removes half of the spread with minimal accuracy gain (base $\to$ SFT: $\Delta$Acc $+0.9$ [$+0.1$, $+1.9$], $\Delta\dr$ $-0.63$ [$-0.74$, $-0.58$]) and the preference-optimization stage restores a little of it ($+0.05$); survey fine-tuning is what buys accuracy. The 9B and 70B DPO runs, trained with the same objective and pair construction, show the same qualitative pattern ($\dr = 0.62$ vs.\ $0.50$), so the collapse is not specific to one family or scale. The fine-tuned model moves toward the conditional mode for each persona, gaining point accuracy at the direct cost of spread.

\subsection{The collapse is not recoverable by raising the sampling temperature}
\label{sec:temp}


We swept the sampling temperature across two model families and scales to test whether the observed collapse is a decoding artifact. If fine-tuning had merely narrowed the output distribution, rather than altered it, raising the temperature should restore the suppressed dispersion. For the most collapsed model in our suite, Tulu~3 70B-DPO (+WVS, $\dr = 0.50$), and for Qwen~3.5 9B-DPO (+WVS), we swept $T \in \{0, 0.5, 0.7, 1.0\}$, from greedy decoding to the untransformed output distribution, with $n = 10{,}000$ at every temperature for both models.

\begin{table}[t]
\centering
\small
\setlength{\tabcolsep}{4pt}
\begin{tabular}{lcccc}
\toprule
\textbf{Metric} & $T = 0$ & $T = 0.5$ & $T = 0.7$ & $T = 1.0$ \\
\midrule
\multicolumn{5}{l}{\textit{Tulu 3 70B-DPO (+WVS)}} \\
Accuracy (\%)      & 57.9 & 58.0 & 57.9 & 57.9 \\
MAE                & 0.78 & 0.78 & 0.78 & 0.78 \\
$\wone$            & 0.55 & 0.54 & 0.55 & 0.54 \\
$\dr$              & 0.50 & 0.50 & 0.50 & 0.51 \\
Preds.\ changed    & ---  & 2.0\% & 2.7\% & 3.7\% \\
\midrule
\multicolumn{5}{l}{\textit{Qwen 3.5 9B-DPO (+WVS)}} \\
Accuracy (\%)      & 53.1 & 53.0 & 52.8 & 52.7 \\
MAE                & 1.04 & 1.05 & 1.05 & 1.06 \\
$\wone$            & 0.73 & 0.73 & 0.73 & 0.73 \\
$\dr$              & 0.62 & 0.62 & 0.63 & 0.65 \\
Preds.\ changed    & ---  & 3.8\% & 5.0\% & 7.1\% \\
\bottomrule
\end{tabular}
\caption{Temperature sweeps on both WVS-fine-tuned DPO models ($n = 10{,}000$ at every temperature). ``Preds.\ changed'' is measured against each model's own $T = 0$ run. The Qwen $T = 0$ column is a fresh greedy run made for the sweep, which is why its accuracy (53.1) differs from Table~\ref{tab:main} (53.0): it differs from the original greedy run on 1.3\% of predictions (126 of 10{,}000), the run-to-run floor of greedy decoding. Even at $T = 1.0$, $\wone$ is unchanged for both models and $\dr$ moves by at most 0.03: sampling perturbs individual predictions without restoring population spread.}
\label{tab:temp}
\end{table}

We find that temperature does not recover the collapse for either model (Table~\ref{tab:temp}). For Tulu, moving from $T = 0$ to $T = 1.0$ changes 3.7\% of predictions and shifts $\dr$ from 0.50 to 0.51. Qwen's output distribution is less concentrated: 7.1\% of its predictions change at $T = 1.0$, roughly twice Tulu's rate, yet $\wone$ stays at 0.73 at every temperature and $\dr$ moves only from 0.62 to 0.65. The pattern holds within WEIRD and non-WEIRD populations separately (Appendix~\ref{app:temp_weird}): Tulu's non-WEIRD $\dr$ stays at 0.33--0.34 across all temperatures. The lost dispersion is therefore not recoverable by sampling temperature; we did not test $T > 1$, nucleus or min-$p$ sampling, or diversity-seeking prompts, so the claim is limited to the range tested. Because DPO compresses variance substantially while SFT and sparse-reward GRPO trained on the same respondent pool barely move it (\S\ref{sec:main_results}), we attribute the loss to the training objective rather than to decoding; this is a behavioural result: we have not measured the output logits of the fine-tuned models directly.


\subsection{Reinforcement learning on Qwen 3.5 9B does not reverse the collapse}
\label{sec:grpo}

We next ask whether reinforcement learning can provide an alternative path. We train Qwen 3.5 9B with GRPO under two reward designs. With a sparse accuracy reward ($+1.0$ for exact match, scaled MAE penalty otherwise), the policy performs near-identically to its zero-shot baseline (38.6\% vs.\ 38.4\% accuracy, $\dr$ 0.70 vs.\ 0.73), a null result (paired $\Delta$Acc $+0.2$ [$-0.2$, $+0.5$]). Human survey data is inherently stochastic, two demographically identical respondents may give different answers, so the sparse reward offers the policy gradient no stable update direction and the model stays at its pre-trained prior. A distribution-shaped reward does not help either: rewarding each sampled answer by its relative frequency among human respondents in the same (country, question) cell, estimated from training questions only, raises accuracy to 47.5\% and lowers $\wone$ to 0.84, but collapses the spread further ($\dr$ 0.70 $\to$ 0.21, reaching 0.00 on Q1). That reward is maximized by always emitting the cell's modal answer, so it drives the policy toward the conditional mean more directly than the sparse reward does.

\subsection{The collapse is most severe for underrepresented populations}
\label{sec:country}

Per-country analysis reveals that consensus collapse is not uniform, and that it is most severe for non-WEIRD populations \citep{henrich2010weirdest}, which are plausibly less represented in pre-training data, although we do not measure that (Table~\ref{tab:country}).

\begin{table}[t]
\centering
\small
\setlength{\tabcolsep}{4pt}
\begin{tabular}{lcc@{\hskip 12pt}lcc}
\toprule
\multicolumn{3}{c}{\textit{WEIRD}} & \multicolumn{3}{c}{\textit{non-WEIRD}} \\
\cmidrule(lr){1-3}\cmidrule(lr){4-6}
\textbf{Ctry} & \textbf{Acc} & $\mathbf{\dr}$ & \textbf{Ctry} & \textbf{Acc} & $\mathbf{\dr}$ \\
\midrule
AUS & 57.5 & 0.79 & BGD & 62.9 & 0.33 \\
DEU & 56.9 & 0.71 & BRA & 51.1 & 0.55 \\
GBR & 53.5 & 0.87 & CHN & 53.9 & 0.60 \\
NLD & 58.9 & 0.70 & ETH & 66.5 & \textbf{0.16} \\
USA & 53.1 & 0.73 & IDN & 68.2 & 0.28 \\
    &      &      & KEN & 51.7 & 0.25 \\
    &      &      & NGA & 60.9 & \textbf{0.11} \\
\bottomrule
\end{tabular}
\caption{Per-country accuracy and $\dr$ for Tulu~3 70B-DPO (+WVS). WEIRD: $\dr = 0.70$--$0.87$. Non-WEIRD collapse: NGA 0.11, ETH 0.16.}
\label{tab:country}
\end{table}

For Nigeria, the aligned model keeps only 11\% of
the human response spread, effectively reducing 660~diverse respondents to
a single stereotyped prediction. This extends the findings of \citet{qu2024bias} along the training trajectory of a single model family (Table~\ref{tab:trajectory}): the pre-trained Llama~3.1 70B base shows no WEIRD/non-WEIRD gap at all ($\dr$ 1.21 vs.\ 1.22, both over-dispersed); the gap opens at the first post-training stage, supervised instruction tuning (Tulu~3 SFT: 0.81 vs.\ 0.43), persists through the DPO and RLVR stages, and deepens further with our survey fine-tuning (0.76 vs.\ 0.33).
\begin{table}[t]
\centering
\small
\setlength{\tabcolsep}{4pt}
\resizebox{\columnwidth}{!}{%
\begin{tabular}{lccc}
\toprule
\textbf{Stage} & \textbf{Acc\%} & $\mathbf{\dr}$ \textbf{WEIRD} & $\mathbf{\dr}$ \textbf{non-WEIRD} \\
\midrule
Llama 3.1 70B (base)  & 50.3 & 1.21 \ci{1.03, 1.37} & 1.22 \ci{1.01, 1.44} \\
Tulu 3 70B-SFT (ZS)   & 51.2 & 0.81 \ci{0.63, 1.01} & 0.43 \ci{0.31, 0.56} \\
Tulu 3 70B-DPO (ZS)   & 49.7 & 0.86 \ci{0.68, 1.05} & 0.49 \ci{0.36, 0.64} \\
Tulu 3 70B-RLVR (ZS)  & 50.6 & 0.83 \ci{0.63, 1.03} & 0.45 \ci{0.33, 0.57} \\
Tulu 3 70B-DPO (+WVS) & 57.9 & 0.76 \ci{0.60, 0.93} & 0.33 \ci{0.22, 0.44} \\
\bottomrule
\end{tabular}}
\caption{WEIRD vs.\ non-WEIRD dispersion retention along the Llama~3.1 70B $\to$ Tulu~3 training trajectory, evaluated on the same 10{,}000 pairs; the first four rows are zero-shot checkpoints, the last is our WVS fine-tune of the Tulu-DPO checkpoint. 95\% cluster-bootstrap CIs over (country, question) cells. The two groups' intervals overlap at the base and separate from the first post-training stage onward.}
\label{tab:trajectory}
\end{table}

\begin{figure}[t]
\centering
\fbox{\includegraphics[width=0.97\columnwidth]{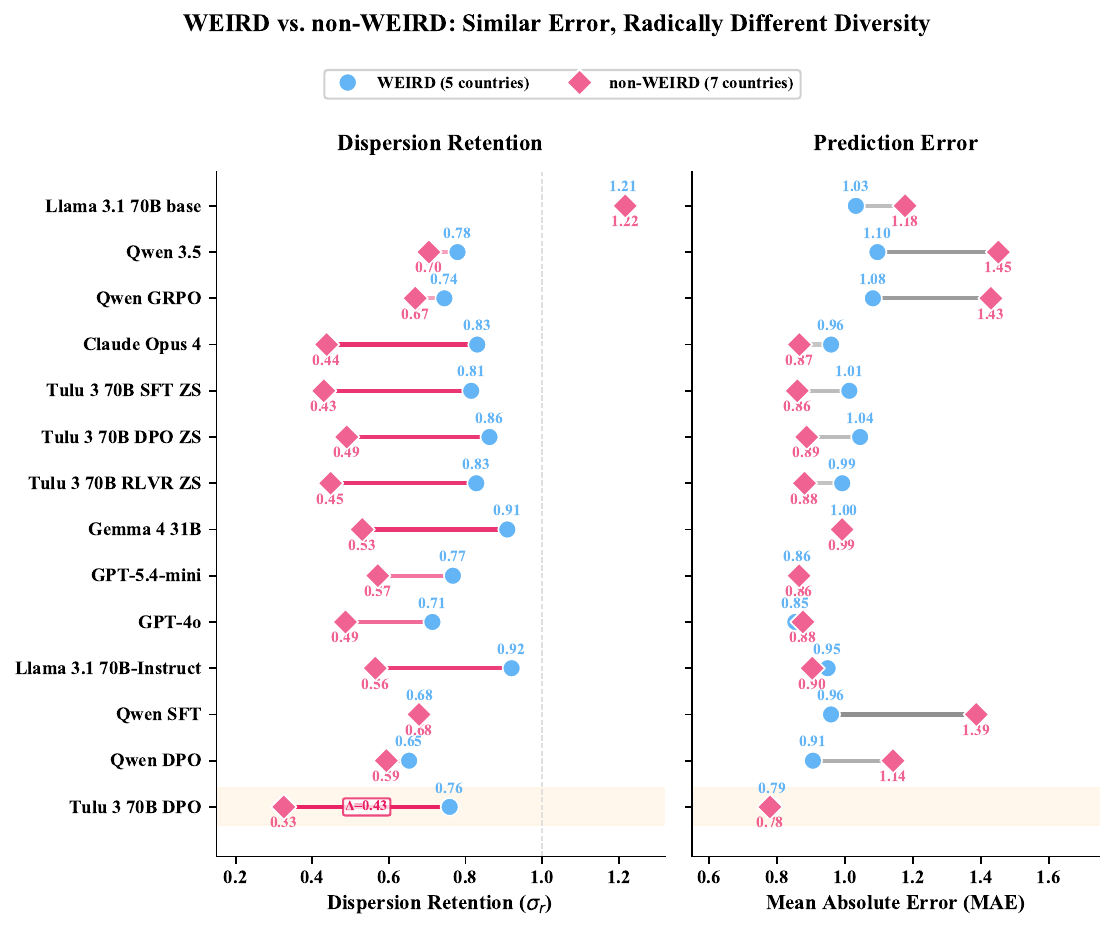}}
\caption{\textbf{WEIRD vs.\ non-WEIRD asymmetry across all models.} Left: dispersion retention ($\dr$); right: prediction error (MAE). Error rates are comparable across populations, but dispersion retention differs by up to $\Delta = 0.43$ for Tulu~3 70B-DPO (+WVS). The pre-trained Llama~3.1 70B base (top row) shows no gap (1.21 vs.\ 1.22). The gap is 0.20--0.43 for every post-trained model outside the Qwen~3.5 9B family, including Tulu's SFT, DPO, and RLVR stages, so the asymmetry is not specific to Tulu or to DPO; within the Qwen family it is at most 0.08 and vanishes after SFT (0.68 vs.\ 0.68).}
\label{fig:weird_gap}
\end{figure}

Figure~\ref{fig:weird_gap} shows that this asymmetry is not unique to Tulu. For every post-trained model outside the Qwen~3.5 9B family, MAE is broadly comparable between WEIRD and non-WEIRD populations (right panel) while dispersion retention diverges by 0.20--0.43 (left panel), with non-WEIRD populations losing more dispersion under every provider's post-training and under our DPO fine-tuning. The Qwen family is the exception: its gap is at most 0.08 zero-shot and after DPO or GRPO, and disappears after SFT (0.68 vs.\ 0.68), so the size of the asymmetry depends on the base model as well as on the training objective.

\subsection{Question-level variation reveals where collapse matters most}
\label{sec:perq}

Figure~\ref{fig:heatmap} shows per-cell $\dr$ across all 12~countries; per-question accuracy and $\dr$ averages are in Appendix~\ref{app:perq}.

\begin{figure}[t]
\fbox{\includegraphics[width=0.97\columnwidth]{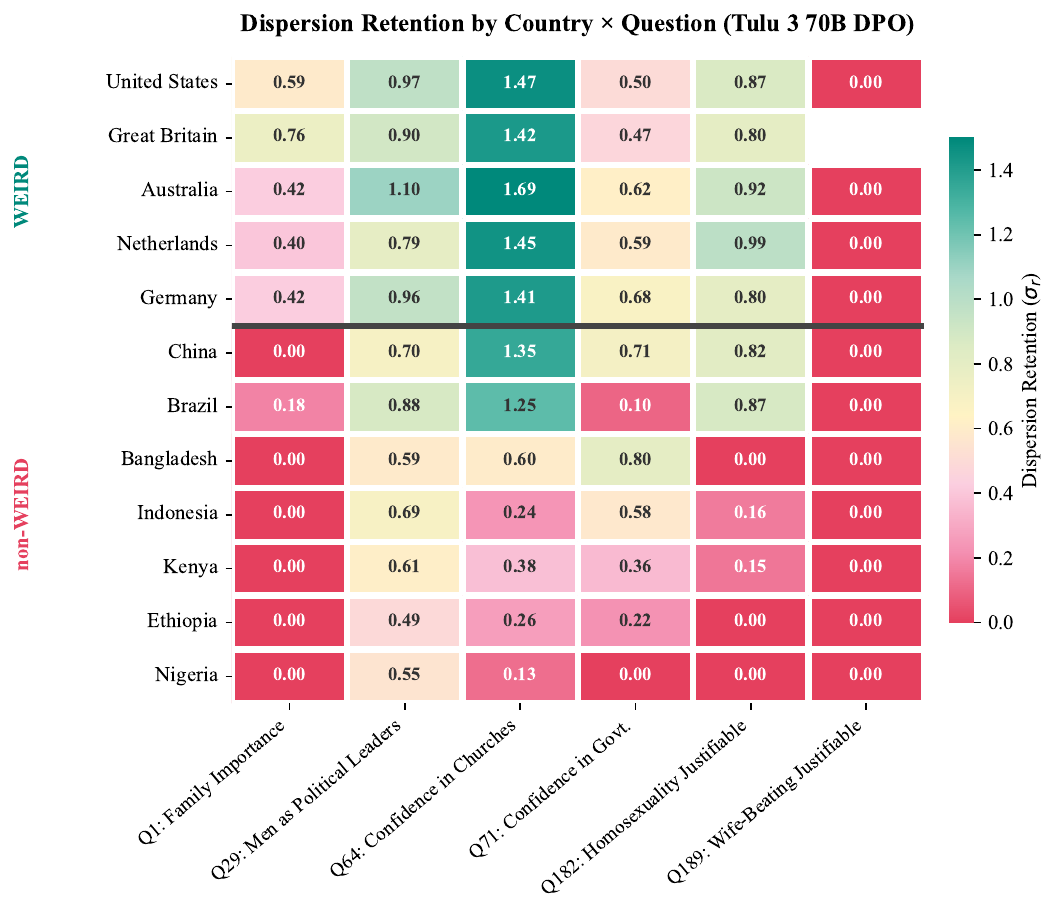}}
\caption{\textbf{Per-cell $\dr$ for Tulu~3 70B-DPO (+WVS)} (12~countries $\times$ 6~questions). The Q189 column is uniformly zero (complete mode collapse). The WEIRD/non-WEIRD divide is visible vertically: WEIRD countries (top rows) retain moderate spread while non-WEIRD countries (bottom rows) show near-total collapse across most questions.}
\label{fig:heatmap}
\end{figure}

\paragraph{Key observations.} For high-consensus questions such as Q1, all models achieve $\sim$88--90\% accuracy, but every model collapses the spread ($\dr < 0.25$); when ground truth is near-unanimous, consensus collapse is harmless. For culturally sensitive questions such as Q182, alignment dominates accuracy (Tulu~3 70B-DPO (+WVS) 52.3\% vs.\ 9.9\% for Qwen~3.5 9B zero-shot), but at the cost of spread ($\dr = 0.53$ vs.\ $0.68$).

\paragraph{Q189: a case study in complete collapse.} Q189 asks: ``\textit{Do you think it is justifiable for a man to beat his wife?}'' (1--10 scale). Tulu~3 70B-DPO (+WVS) predicts 1 (``never justifiable'') for every single respondent in every country\footnote{Q189 was not administered in Great Britain in the WVS Wave~7; the 1{,}544 predictions span the remaining 11~countries.}---1{,}544 identical predictions, $\dr = 0.00$. Yet the human distribution varies substantially: while the global mode is~1, Kenya's mean response is 3.44 ($\sigma = 2.95$), Nigeria's is 2.23 ($\sigma = 1.99$), and Bangladesh's is 2.31 ($\sigma = 1.73$). The model has learned the socially desirable answer, the one DPO's preference signal reinforces, and applies it uniformly, erasing all cross-national and within-country variation. GPT-4o exhibits identical behavior ($\dr = 0.00$), suggesting this is not specific to fine-tuning but reflects a broader tendency of aligned models to converge on normatively ``correct'' responses to morally charged questions.

\subsection{How Much Can a Post-Hoc Fix Recover?}
\label{sec:mitigation}

To test whether the collapse is recoverable by post-hoc means, we ask how much of the spread can be restored through a minimal intervention, which we call \textit{prior mixing}: probabilistically mixing the aligned model with a small unaligned prior from a different family, Qwen~3.5 9B zero-shot. This is not proposed as a deployment-ready method, but as a probe of the collapse's depth---a test of what mixing can and cannot recover.

\paragraph{Probe Formulation.} Let $M_{\text{DPO}}$ denote the aligned point-predictor and $M_{\text{ZS}}$ the unaligned zero-shot prior. For each simulated respondent $i$, sample $z_i \sim \text{Bernoulli}(\alpha)$:
\begin{equation}
\hat{y}_i = z_i \cdot M_{\text{DPO}}(x_i) + (1 - z_i) \cdot M_{\text{ZS}}(x_i)
\label{eq:ensemble}
\end{equation}
At $\alpha = 1.0$ the probe reduces to pure DPO and at $\alpha = 0.0$ to the prior. Each query is served by exactly one model, so with a 9B prior at $1-\alpha = 0.05$ the mixture is cheaper than the 70B model alone, and unlike rectification \citep{krsteski2025valid} it needs no ground-truth human data at inference time.

\paragraph{Protocol.} We partition the 10{,}000 test pairs into a 20\% development set (1{,}995 pairs) and an 80\% held-out set (8{,}005 pairs), stratified by (country, question) cell with a fixed seed, select $\alpha$ on the development set, and report the probe on the held-out set; the diagnostic results in \S\ref{sec:results} use the full set, on which no selection was performed. Sweeping $\alpha$ on the development set (Appendix~\ref{app:mixing}), we select $\alpha = 0.95$, the largest value whose dev-set $\wone$ (0.58) is no worse than pure DPO's (0.60) while $\dr$ improves substantially (0.60 vs.\ 0.52).

\begin{table}[t]
\centering
\small
\resizebox{\columnwidth}{!}{%
\begin{tabular}{lcccc}
\toprule
\textbf{Configuration} & \textbf{Acc\%} & \textbf{MAE} & $\mathbf{\wone}$ $\downarrow$ & $\mathbf{\dr}$ ($\rightarrow 1$) \\
\midrule
DPO only ($\alpha = 1.0$) & 58.3 & 0.77 & 0.54 & 0.51 \\
Zero-shot prior only ($\alpha = 0.0$) & 39.0 & 1.28 & 1.00 & 0.74 \\
\midrule
\textbf{Prior mixing ($\alpha = 0.95$)} & \textbf{57.0} & \textbf{0.80} & \textbf{0.51} & \textbf{0.62} \\
\bottomrule
\end{tabular}}
\caption{Held-out test results ($n = 8{,}005$) using the dev-selected $\alpha = 0.95$ and routing seed 42; across ten routing seeds the mixed row gives $\wone = 0.51 \pm 0.01$ and $\dr = 0.60 \pm 0.02$. Prior mixing reduces $\wone$ from 0.54 to 0.51 and partially recovers $\dr$ from 0.51 to 0.62. The recovery is consistent but bounded; the residual gap measures how much of the collapse persists under the simplest post-hoc intervention, particularly for non-WEIRD populations (Table~\ref{tab:country_ens}).}
\label{tab:test_results}
\end{table}

\begin{table}[t]
\centering
\small
\begin{tabular}{lcc|cc}
\toprule
 & \multicolumn{2}{c}{\textbf{Accuracy (\%)}} & \multicolumn{2}{c}{$\mathbf{\dr}$} \\
\cmidrule(lr){2-3}\cmidrule(lr){4-5}
\textbf{Country} & \textbf{DPO} & \textbf{Mix.} & \textbf{DPO} & \textbf{Mix.} \\
\midrule
AUS & 58.2 & 56.9 & 0.81 & 0.90 \\
DEU & 56.4 & 55.1 & 0.70 & 0.75 \\
GBR & 54.6 & 53.7 & 0.85 & 0.87 \\
NLD & 59.8 & 58.7 & 0.71 & 0.75 \\
USA & 54.4 & 54.0 & 0.74 & 0.76 \\
\midrule
BGD & 61.9 & 60.9 & 0.30 & 0.51 \\
BRA & 50.7 & 48.9 & 0.56 & 0.73 \\
CHN & 53.6 & 52.7 & 0.59 & 0.64 \\
ETH & 67.5 & 66.7 & 0.18 & 0.26 \\
IDN & 68.7 & 66.4 & 0.29 & 0.53 \\
KEN & 50.7 & 49.0 & 0.25 & 0.37 \\
NGA & 62.4 & 61.1 & 0.13 & 0.36 \\
\bottomrule
\end{tabular}
\caption{Per-country recoverability under prior mixing ($\alpha = 0.95$) on the held-out test set ($n = 8{,}005$). The residual gap from $\dr = 1.0$ measures the structural component of the collapse, i.e., spread not recoverable by routing a fraction of queries to an unaligned prior. For non-WEIRD populations, large residuals remain: Nigeria recovers to $\dr = 0.36$, meaning 64\% of the within-country spread among Nigerian respondents is not recoverable by output-space mixing.}
\label{tab:country_ens}
\end{table}

Prior mixing raises held-out $\dr$ from 0.51 to 0.62 and lowers $\wone$ from 0.54 to 0.51 (Table~\ref{tab:test_results}), a small but consistent improvement over pure DPO; held-out $\wone$ is flat at 0.50--0.51 across $\alpha \in [0.80, 0.95]$, so the recovery is not an artifact of a brittle $\alpha$ choice. Decomposed by country (Table~\ref{tab:country_ens}), the bounded recovery is itself the finding. For WEIRD populations, whose retention was already 0.70--0.85, mixing adds a few points (AUS $0.81 \to 0.90$, DEU $0.70 \to 0.75$). For non-WEIRD populations the gap remains large even after mixing (NGA $0.13 \to 0.36$, IDN $0.29 \to 0.53$, KEN $0.25 \to 0.37$), so the unaligned prior does not supply the lost spread for these populations either. Evidence from these results suggests that collapse for underrepresented populations is not a surface-level reweighting failure but a deeper representational gap that output-space mixing with an unaligned prior cannot recover.

\section{Discussion}
\label{sec:discussion}

\subsection{Why value-encoded personas are not enough}

Consensus collapse reflects, in part, a gap between categorical persona descriptions and individual biographical specificity. Our value-encoded personas are not simple demographic vectors, but instead each encodes $\sim$60 words of information, including interpersonal trust orientation, civic engagement level, and religious identity. Crucially, they encode the Inglehart--Welzel cultural-map dimensions (traditional/secular-rational, survival/self-expression) that the WVS instrument was specifically designed to measure. Despite this, the model still reduces Nigeria to $\dr = 0.11$, keeping only 11\% of the spread that exists among Nigerian respondents who share the same categorical profile. The failure does not appear to be caused by lack of information, since our personas already encode the most meaningful dimensions of the WVS instrument. We believe that part of the deficit therefore lies in the lack of \textit{idiosyncratic biographical detail} required for individual-level prediction, but not all of it. A Bayes-optimal point predictor given only the persona would retain $\dr \approx 0.31$ on average (Appendix~\ref{app:oracle}), and Nigeria, Ethiopia, and Kenya fall below even that floor, so the model discards variance the persona demonstrably supports, while Western countries sit about twice above it. Two respondents who are both ``traditional, trusting, civically embedded, 40-year-old Orthodox males from Ethiopia'' may hold fundamentally different views shaped by life experiences not captured at the persona level.

Formally, under cross-entropy loss with input features $X$ that do not fully determine target $Y$, the loss-minimizing predictor converges to $\mathbb{E}[Y \mid X]$, whose variance satisfies $\text{Var}(\mathbb{E}[Y \mid X]) \leq \text{Var}(Y)$, with equality only when $X$ fully determines $Y$. This bounds the variance any point predictor can express when the input does not resolve individual responses, and it is the direction of the compression we observe; but the bound alone does not explain the pattern, since the fine-tuned model sits above the floor for WEIRD countries and below it for the most affected non-WEIRD countries (Appendix~\ref{app:oracle}). \citet{park2024generative} provide compelling evidence for this: replacing structured personas with two-hour interview transcripts that inherently capture biographical specificity enables more detailed modeling, with agents reproducing participants' survey answers 85\% as accurately as the participants reproduce their own answers two weeks later. These findings suggest a continuum of input richness along which consensus collapse may progressively dissolve, scaling from bare demographics through the structured personas evaluated in this work, and ultimately to dense biographical contexts \citep{park2024generative}.

\subsection{Implications for the field}

Our diagnostic findings carry three implications. First, point accuracy alone can be a misleading metric. The most accurate model in our study is also the least distributionally faithful, and researchers evaluating SRM systems must report variance-aware metrics ($\wone$, $\dr$) alongside accuracy. Second, alignment exacerbates representational harm, where the populations most in need of accurate simulation, underrepresented and non-Western communities, suffer the most substantial collapse, plausibly a consequence of alignment objectives applied to populations less represented in the base model training data, though we do not measure coverage directly. Third, decoding settings do not appear to reverse the collapse, so recovering spread requires an intervention beyond temperature. This could include richer per-individual context \citep{park2024generative}, post-hoc strategies such as rectification or prior mixing, or prompt-level distribution elicitation, which restored spread for GPT-4o at a substantial accuracy cost (Appendix~\ref{app:prompt_probe}).

\section{Conclusion}
\label{sec:conclusion}

In this work, we identified a failure mode termed \textit{consensus collapse}, which is the systematic compression of response spread induced by post-training and survey fine-tuning. We demonstrate that general post-training removes half of the spread with minimal accuracy gain; survey fine-tuning then buys accuracy at a further cost, so that the most accurate model (Tulu~3 70B-DPO fine-tuned on WVS, 57.9\%) keeps only half of the human spread ($\dr = 0.50$). Sampling at up to $T = 1.0$ leaves $\wone$ unchanged for both fine-tuned DPO models, and reinforcement learning on Qwen~3.5 9B left the model at its pre-trained prior under a sparse accuracy reward and collapsed variance further under a distribution-shaped one. We also find that the collapse is inequitable, with WEIRD populations keeping 70--87\% of their spread while Nigeria keeps 11\%, and on Q189 the aligned model predicted ``never justifiable'' for all 1{,}544 respondents despite a Kenyan mean of 3.44 on a 10-point scale. Prior mixing, the simplest post-hoc probe, lowers $\wone$ from 0.54 to 0.51 but lifts Nigeria only from $\dr = 0.13$ to $0.36$; the bounded recovery shows that the collapse is not stored where output-space mixing can reach it. Work that simulates populations with LLMs must therefore move beyond point accuracy and recognize that current post-training and survey fine-tuning trade population diversity for consensus.

\section*{Limitations}

Our evaluation uses 6~held-out WVS questions from a single survey instrument. We note three mitigating factors: (1)~we hypothesize, but have not tested, that the underlying mechanism (cross-entropy loss with features that do not fully determine the target, \S\ref{sec:discussion}) is instrument-agnostic; (2)~the WVS is among the most rigorously cross-validated cross-cultural instruments, with the Inglehart--Welzel dimensions encoded directly into our personas; and (3)~the per-country asymmetry we document (e.g., UK $\dr = 0.87$, Netherlands $0.70$ vs.\ Nigeria $\dr = 0.11$, Ethiopia $0.16$) mirrors patterns \citet{durmus2024measuring} observe across different cultural instruments. Nonetheless, replication on other instruments (Pew, ANES, Eurobarometer) is valuable future work. 

The (Q189, GBR) cell is empty across all data files due to a gap in the WVS source data, yielding 71/72 cell coverage (98.6\%). The persona's two value tiers are derived from the WVS SACSECVAL and RESEMAVAL indices, which are computed from multiple survey items; two of our test questions (Q29 and Q182) and two training questions (Q6 and Q185) are among them. Excluding the two test items leaves the accuracy--diversity ordering unchanged (Appendix~\ref{app:leakage}); we did not rebuild the personas with these items removed. All prompts were administered in English, whereas WVS respondents in non-English-speaking countries answered in local languages. Our decoding sweep covers sampling temperatures up to 1.0 only; higher temperatures, nucleus or min-$p$ sampling, and diversity-seeking prompts were not tested on the fine-tuned models. A prompt-level probe on GPT-4o (Appendix~\ref{app:prompt_probe}) shows that eliciting a stated answer distribution and sampling from it restores spread on the collapsed item Q189 at a cost of 12 accuracy points; we have not run the probe on the fine-tuned models. Our evaluation covers 4-point and 10-point ordinal scales; generalization to wider scales or open-ended responses requires further investigation. The prior-mixing $\alpha$ was selected on a stratified 20\% development split; using a larger or independently collected development set would further strengthen the claim. Furthermore, our evaluation relies on a standardized, neutral persona prompt. While this isolates the effect of the model's training, investigating whether prompt engineering can recover the spread of the fine-tuned models remains an important direction for future work.

\section*{Ethical Considerations}

All data is from the publicly available World Values Survey (WVS), a well-established cross-national survey instrument with documented sampling protocols and methodology used across the social sciences. Our use of the WVS complies with its public release terms.

All model usage adhered to the respective provider terms of service and API usage policies for proprietary models (OpenAI, Anthropic, Google), and to the model license terms for open-weight models (Qwen, Gemma, Llama, Tulu) accessed via Hugging Face and OpenRouter. No usage policies were circumvented at any stage of evaluation or fine-tuning.

We evaluate simulation fidelity rather than advocate for replacing human participants. Simulated survey responses should never substitute for genuine engagement with real communities, particularly marginalized populations. The geographic labels we employ are analytical shorthand and should not be read as essentializing any national identity.

\bibliography{references}
\appendix
\begin{table*}[t]
\centering
\small
\caption{Complete question inventory drawn from WVS Wave~7. Q57 (interpersonal trust) is used in persona construction and excluded from evaluation. $^{\dagger}$Q29 and Q182 are components of the RESEMAVAL index encoded in the persona, as are the training items Q6 and Q185 (Appendix~\ref{app:leakage}).}
\label{tab:questions}
\begin{tabular}{llccc}
\toprule
\textbf{Code} & \textbf{Question (abbreviated)} & \textbf{Scale} & \textbf{Cluster} & \textbf{Split} \\
\midrule
Q1   & How important is family in your life?                          & 1--4  & Values       & Test  \\
Q2   & How important are friends in your life?                       & 1--4  & Values       & Train \\
Q6   & How important is religion in your life?                       & 1--4  & Values       & Train \\
Q27  & Main goal: make my parents proud                              & 1--4  & Values       & Train \\
Q29  & Men make better political leaders than women                  & 1--4  & Values       & Test$^{\dagger}$ \\
Q46  & Would you say you are happy?                                  & 1--4  & Economic     & Train \\
Q47  & State of health these days                                    & 1--5  & Economic     & Train \\
Q48  & Freedom of choice and control over your life                  & 1--10 & Economic     & Train \\
Q49  & Satisfaction with life as a whole                             & 1--10 & Economic     & Train \\
Q50  & Satisfaction with financial situation of household            & 1--10 & Economic     & Train \\
Q58  & How much do you trust your family?                            & 1--4  & Social Trust & Train \\
Q59  & How much do you trust people in your neighborhood?            & 1--4  & Social Trust & Train \\
Q60  & How much do you trust people you know personally?             & 1--4  & Social Trust & Train \\
Q64  & Confidence in churches/mosques/temples                        & 1--4  & Social Trust & Test  \\
Q71  & Confidence in the government                                  & 1--4  & Political    & Test  \\
Q106 & Incomes equal vs.\ incentives for individual effort           & 1--10 & Economic     & Train \\
Q164 & How important is God in your life?                            & 1--10 & Values       & Train \\
Q182 & Is homosexuality justifiable?                                 & 1--10 & Values       & Test$^{\dagger}$ \\
Q185 & Is divorce justifiable?                                       & 1--10 & Values       & Train \\
Q189 & Is it justifiable for a man to beat his wife?                 & 1--10 & Values       & Test  \\
Q240 & Political self-placement: left--right                         & 1--10 & Political    & Train \\
\bottomrule
\end{tabular}
\end{table*}

\section{Question Inventory}
\label{app:questions}
Table~\ref{tab:questions} lists all 21 World Values Survey questions used in this study, their response scales, thematic clusters, and train/test assignment. The test set spans three of the four clusters (Values, Social Trust, Political); no Economic item is held out, so generalisation to that cluster is not tested.

\section{Persona--Index Overlap}
\label{app:leakage}
The two value tiers in each persona are three-level binnings of the WVS SACSECVAL and RESEMAVAL indices, which the WVS computes from twelve survey items each. Following the Common EVS/WVS Dictionary, two test questions are among these items, Q29 (D059, the gender-equality component of RESEMAVAL) and Q182 (F118, the choice component of RESEMAVAL), as are two training questions, Q6 (A006, SACSECVAL) and Q185 (F121, RESEMAVAL). The remaining four test questions are not part of either index. Table~\ref{tab:leakage} measures how much signal the self-expression tier carries for the two affected test items, with Q189 as a control, and recomputes the headline metrics on the four unaffected test questions.

\begin{table}[h]
\centering
\small
\begin{tabular}{lccc}
\toprule
\textbf{Q} & \textbf{In index} & \textbf{Tier adds} & \textbf{corr(tier, answer)} \\
\midrule
Q182 & yes & $+1.7$ pts & $+0.42$ \\
Q29  & yes & $+7.0$ pts & $+0.33$ \\
Q189 & no (control) & $+0.1$ pts & $+0.09$ \\
\bottomrule
\end{tabular}
\vspace{4pt}

\resizebox{\columnwidth}{!}{%
\begin{tabular}{lcccc}
\toprule
& \multicolumn{2}{c}{\textbf{Acc (\%)}} & \multicolumn{2}{c}{$\dr$} \\
\cmidrule(lr){2-3}\cmidrule(lr){4-5}
\textbf{Model} & all 6 & clean 4 & all 6 & clean 4 \\
\midrule
Tulu 3 70B-DPO (+WVS) & 57.9 & 64.2 & 0.50 & 0.43 \\
Tulu 3 70B-DPO (ZS)   & 49.7 & 58.7 & 0.64 & 0.50 \\
Qwen 3.5 9B-DPO (+WVS) & 53.0 & 59.6 & 0.62 & 0.64 \\
Qwen 3.5 9B (ZS)      & 38.4 & 47.5 & 0.73 & 0.73 \\
GPT-4o (ZS)           & 52.8 & 61.4 & 0.58 & 0.41 \\
\bottomrule
\end{tabular}}
\caption{Persona--index overlap. \emph{Top:} accuracy gained by predicting the modal answer per (country, persona-tier) cell over a per-country modal baseline, and the mean within-country correlation between the three-level self-expression tier and the ordinal answer, computed on ground-truth responses only; Q189 is not part of the index and serves as a control. \emph{Bottom:} accuracy and $\dr$ recomputed on the four test questions outside the index (Q1, Q64, Q71, Q189). The accuracy--diversity ordering is unchanged.}
\label{tab:leakage}
\end{table}

\section{Bayes-Optimal Variance Floor}
\label{app:oracle}
Because personas do not fully determine responses, $\dr = 1$ is unattainable for any point predictor. We estimate the attainable floor from the full WVS microdata (462K records): respondents are grouped into identical-persona cells (the persona text with age replaced by its decade), and for each (country, test question) pair the explained variance is estimated by the law of total variance with unbiased within-cell variances, $\mathrm{Var}(\mathbb{E}[Y \mid X]) = \mathrm{Var}(Y) - \mathbb{E}[\mathrm{Var}(Y \mid X)]$, dropping cells with fewer than five respondents. The oracle $\dr$ is the square root of the explained fraction; it averages 0.31 (per question: Q1 0.08, Q29 0.36, Q64 0.34, Q71 0.30, Q182 0.47, Q189 0.29). Table~\ref{tab:oracle} compares it with Tulu~3 70B-DPO (+WVS).

\begin{table}[h]
\centering
\small
\begin{tabular}{lccc}
\toprule
\textbf{Country} & \textbf{Oracle} $\dr$ & \textbf{Tulu +WVS-DPO} & \textbf{Ratio} \\
\midrule
AUS & 0.35 & 0.79 & 2.26 \\
DEU & 0.30 & 0.71 & 2.37 \\
GBR & 0.35 & 0.87 & 2.48 \\
NLD & 0.29 & 0.70 & 2.42 \\
USA & 0.41 & 0.73 & 1.77 \\
\midrule
BGD & 0.24 & 0.33 & 1.36 \\
BRA & 0.20 & 0.55 & 2.71 \\
CHN & 0.26 & 0.60 & 2.28 \\
ETH & 0.38 & 0.16 & \textbf{0.43} \\
IDN & 0.26 & 0.28 & 1.08 \\
KEN & 0.33 & 0.25 & \textbf{0.77} \\
NGA & 0.31 & 0.11 & \textbf{0.35} \\
\bottomrule
\end{tabular}
\caption{Oracle deviation ratio of a Bayes-optimal point predictor that outputs $\mathbb{E}[Y \mid \text{persona}]$, against the fine-tuned model. Ratio $=$ model $\dr$ / oracle $\dr$. Western countries sit about twice \emph{above} the floor, whereas Ethiopia, Kenya, and Nigeria fall \emph{below} it: the model under-expresses even the variance its input supports.}
\label{tab:oracle}
\end{table}

\section{Temperature Analysis by Subgroup}
\label{app:temp_weird}
Tables~\ref{tab:temp_weird} and~\ref{tab:temp_weird_qwen} disaggregate the temperature sweeps of \S\ref{sec:temp} by WEIRD and non-WEIRD populations. For Tulu~3 70B-DPO (+WVS) the gap between the groups is constant across temperatures: non-WEIRD $\dr$ remains at 0.33--0.34 from $T=0$ to $T=1.0$, while WEIRD $\dr$ holds at 0.76. For Qwen~3.5 9B-DPO, whose predictions change at roughly twice Tulu's rate, $\wone$ is flat in both groups; non-WEIRD $\dr$ drifts from 0.59 to 0.64 without any accompanying $\wone$ improvement, which we read as added noise rather than recovered signal. Stochastic sampling does not reach the suppressed variance for either group, and reaches it least where the collapse is most severe.

\begin{table}[h]
\centering
\small
\resizebox{\columnwidth}{!}{%
\begin{tabular}{llcccc}
\toprule
\textbf{Group} & \textbf{Metric} & $T = 0$ & $T = 0.5$ & $T = 0.7$ & $T = 1.0$ \\
\midrule
\multirow{4}{*}{WEIRD}
 & Accuracy (\%)   & 55.8 & 55.9 & 55.9 & 55.8 \\
 & $\wone$         & 0.40 & 0.39 & 0.40 & 0.39 \\
 & $\dr$           & 0.76 & 0.76 & 0.76 & 0.76 \\
 & Preds.\ changed & ---  & 2.1\% & 3.4\% & 4.4\% \\
\midrule
\multirow{4}{*}{Non-WEIRD}
 & Accuracy (\%)   & 59.5 & 59.6 & 59.4 & 59.5 \\
 & $\wone$         & 0.65 & 0.65 & 0.65 & 0.65 \\
 & $\dr$           & 0.33 & 0.33 & 0.33 & 0.34 \\
 & Preds.\ changed & ---  & 1.9\% & 2.2\% & 3.2\% \\
\bottomrule
\end{tabular}}
\caption{Temperature sweep on Tulu~3 70B-DPO (+WVS) disaggregated by WEIRD and non-WEIRD populations. $\dr$ for non-WEIRD countries remains at 0.33--0.34 across all temperatures; the gap between the groups is unchanged by sampling.}
\label{tab:temp_weird}
\end{table}

\begin{table}[h]
\centering
\small
\resizebox{\columnwidth}{!}{%
\begin{tabular}{llcccc}
\toprule
\textbf{Group} & \textbf{Metric} & $T = 0$ & $T = 0.5$ & $T = 0.7$ & $T = 1.0$ \\
\midrule
\multirow{4}{*}{WEIRD}
 & Accuracy (\%)   & 51.6 & 51.1 & 50.7 & 51.2 \\
 & $\wone$         & 0.54 & 0.56 & 0.56 & 0.54 \\
 & $\dr$           & 0.65 & 0.65 & 0.66 & 0.66 \\
 & Preds.\ changed & ---  & 3.4\% & 4.7\% & 5.9\% \\
\midrule
\multirow{4}{*}{Non-WEIRD}
 & Accuracy (\%)   & 54.2 & 54.3 & 54.4 & 53.8 \\
 & $\wone$         & 0.85 & 0.85 & 0.84 & 0.86 \\
 & $\dr$           & 0.59 & 0.60 & 0.61 & 0.64 \\
 & Preds.\ changed & ---  & 4.1\% & 5.2\% & 8.0\% \\
\bottomrule
\end{tabular}}
\caption{Temperature sweep on Qwen~3.5 9B-DPO (+WVS) disaggregated by WEIRD and non-WEIRD populations. Predictions change at roughly twice Tulu's rate, yet $\wone$ is flat in both groups; the non-WEIRD $\dr$ drift from 0.59 to 0.64 is not accompanied by any $\wone$ improvement.}
\label{tab:temp_weird_qwen}
\end{table}

\section{Prompt-Level Remedies on GPT-4o}
\label{app:prompt_probe}
To test whether a single closed-weight model can be prompted out of the collapse without a second checkpoint, we ran three conditions on GPT-4o over the same stratified subset of the evaluation set: 30 respondents per (country, question) cell, 2{,}130 pairs in total (seed 3407), with no parse failures in any condition. \emph{Baseline} is the paper's prompt decoded greedily. \emph{Population draw} appends an instruction to answer as one randomly drawn individual from the real-world population matching the profile rather than giving the most typical or most socially acceptable answer, and samples at $T = 1.0$. \emph{Distribution elicitation} asks the model to return a JSON probability distribution over the answer scale (decoded greedily) and samples one answer per respondent from that stated distribution with a per-respondent seed. Table~\ref{tab:prompt_probe} reports the results.

\begin{table}[h]
\centering
\small
\resizebox{\columnwidth}{!}{%
\begin{tabular}{lcccccc}
\toprule
\textbf{Condition} & \textbf{Acc\%} & \textbf{MAE} & $\mathbf{\wone}$ $\downarrow$ & $\mathbf{\dr}$ & $\mathbf{\dr}$ \textbf{WEIRD} & $\mathbf{\dr}$ \textbf{non-WEIRD} \\
\midrule
Baseline                 & 51.2 & 0.88 & 0.63 & 0.61 & 0.77 & 0.49 \\
Population draw          & 45.2 & 0.98 & 0.63 & 0.68 & 0.76 & 0.63 \\
Distribution elicitation & 39.5 & 1.19 & 0.57 & 1.34 & 1.32 & 1.35 \\
\bottomrule
\end{tabular}}
\vspace{4pt}

\begin{tabular}{lccc}
\toprule
\textbf{Q} & \textbf{Baseline} & \textbf{Population} & \textbf{Distribution} \\
\midrule
Q1   & 0.00 & 0.10 & 2.02 \\
Q29  & 1.08 & 0.94 & 1.36 \\
Q64  & 1.10 & 1.14 & 1.46 \\
Q71  & 0.48 & 0.60 & 1.19 \\
Q182 & 0.79 & 0.76 & 1.13 \\
Q189 & 0.00 & 0.37 & 1.02 \\
\bottomrule
\end{tabular}
\caption{Prompt-level remedies on GPT-4o ($n = 2{,}130$, 30 pairs per cell, identical respondents in every condition). \emph{Top:} aggregate metrics. \emph{Bottom:} per-question $\dr$. Population-draw prompting recovers little spread (0.61 $\to$ 0.68, mostly on Q189) and costs 6 accuracy points. Distribution elicitation restores the collapsed item Q189 ($\dr$ 0.00 $\to$ 1.02) and lowers $\wone$ to 0.57, comparable to prior mixing without a second model, but over-disperses consensus items (Q1: $\dr = 2.02$) and costs 12 accuracy points. The probe was run on GPT-4o only; it does not test whether prompt-level elicitation reaches the spread erased by survey fine-tuning.}
\label{tab:prompt_probe}
\end{table}

\section{Prior-Mixing $\alpha$ Sweep}
\label{app:mixing}
Table~\ref{tab:sweep} reports the development-set $\alpha$ sweep behind the selection rule in \S\ref{sec:mitigation}.

\begin{table}[h]
\centering
\small
\begin{tabular}{ccccc}
\toprule
$\alpha$ & \textbf{Acc\%} & \textbf{MAE} & $\mathbf{\wone}$ $\downarrow$ & $\mathbf{\dr}$ ($\rightarrow 1$) \\
\midrule
0.00 & 36.4 & 1.36 & 1.07 & 0.97 \\
\midrule
0.50 & 45.9 & 1.10 & 0.70 & 0.98 \\
0.60 & 47.8 & 1.03 & 0.63 & 0.97 \\
0.70 & 50.6 & 0.98 & 0.59 & 0.94 \\
0.80 & 53.1 & 0.92 & 0.55 & 0.82 \\
0.85 & 53.8 & 0.91 & 0.55 & 0.78 \\
0.90 & 54.7 & 0.88 & 0.55 & 0.72 \\
\textbf{0.95} & \textbf{55.6} & \textbf{0.85} & \textbf{0.58} & \textbf{0.60} \\
\midrule
1.00 & 56.5 & 0.82 & 0.60 & 0.52 \\
\bottomrule
\end{tabular}
\caption{$\alpha$ sweep on the development set ($n = 1{,}995$; routing seed 42). $\wone$ falls from 1.07 at $\alpha = 0$ to 0.58 at $\alpha = 0.95$, against 0.60 for pure DPO. We select $\alpha = 0.95$, the largest $\alpha$ whose dev-set $\wone$ is no worse than pure DPO's while $\dr$ improves substantially (0.60 vs.\ 0.52).}
\label{tab:sweep}
\end{table}

\section{Per-Question Accuracy and Dispersion Retention}
\label{app:perq}

Tables~\ref{tab:perq_acc} and~\ref{tab:perq_dr} report per-question accuracy and deviation ratio averaged across all 12~countries, complementing the per-cell breakdown in Figure~\ref{fig:heatmap}. For high-consensus questions such as Q1, all models reach $\sim$88--90\% accuracy but collapse the spread ($\dr < 0.25$): when ground truth is near-unanimous, collapse is harmless. For culturally sensitive questions, alignment dominates accuracy at the cost of spread --- on Q182, Tulu~3 70B-DPO (+WVS) reaches 52.3\% (vs.\ 9.9\% for Qwen~3.5 9B zero-shot and 33.0\% for GPT-4o) but at $\dr = 0.53$ (vs.\ $0.68$ for Qwen zero-shot). On Q189 (``justifiable for a man to beat his wife''), Tulu~3 70B-DPO (+WVS) and GPT-4o both achieve $\dr = 0.00$, complete mode collapse on the same morally charged item, while unaligned models retain $\dr > 1.0$. On Q64 (confidence in religious institutions), unaligned models over-diversify ($\dr > 1.0$), suggesting that some of the spread they preserve is noise rather than signal.

\begin{table}[h]
\centering
\small
\resizebox{\columnwidth}{!}{
\begin{tabular}{lcccccc}
\toprule
\textbf{Q} & \textbf{Tulu} & \textbf{Qw.DPO} & \textbf{Qw.SFT} & \textbf{GPT-4o} & \textbf{Qw.ZS} & \textbf{GRPO} \\
\midrule
Q1   & 88.0 & 88.4 & 89.9 & 89.8 & 89.5 & 89.7 \\
Q29  & 39.5 & 32.8 & 24.1 & 39.1 & 31.9 & 32.2 \\
Q64  & 49.2 & 50.7 & 51.4 & 43.4 & 47.5 & 48.0 \\
Q71  & 42.1 & 37.3 & 34.4 & 35.2 & 32.6 & 34.9 \\
Q182 & 52.3 & 47.6 & 60.6 & 33.0 &  9.9 &  6.8 \\
Q189 & 79.0 & 62.8 & 58.4 & 79.0 & 18.5 & 19.2 \\
\bottomrule
\end{tabular}
}
\caption{Per-question accuracy (\%). Q1 is high-consensus across all models. Q182 and Q189 show the largest spread across models: Qwen~3.5 9B zero-shot and sparse-reward GRPO score 7--19\%, while the WVS-fine-tuned models and GPT-4o score 33--79\%.}
\label{tab:perq_acc}
\end{table}

\begin{table}[h]
\centering
\small
\resizebox{\columnwidth}{!}{
\begin{tabular}{lcccccc}
\toprule
\textbf{Q} & \textbf{Tulu} & \textbf{Qw.DPO} & \textbf{Qw.SFT} & \textbf{GPT-4o} & \textbf{Qw.ZS} & \textbf{GRPO} \\
\midrule
Q1   & 0.23 & 0.22 & 0.00 & 0.03 & 0.10 & 0.05 \\
Q29  & 0.77 & 0.54 & 0.75 & 1.00 & 0.80 & 0.73 \\
Q64  & 0.97 & 0.89 & 0.73 & 1.09 & 1.17 & 1.11 \\
Q71  & 0.47 & 0.47 & 0.59 & 0.49 & 0.50 & 0.48 \\
Q182 & 0.53 & 0.60 & 0.75 & 0.82 & 0.68 & 0.67 \\
Q189 & 0.00 & 1.02 & 1.32 & 0.00 & 1.20 & 1.20 \\
\bottomrule
\end{tabular}
}
\caption{Per-question deviation ratio ($\dr$), averaged across all 12~countries. Complete collapse on Q189 ($\dr = 0.00$) for both Tulu~3 70B-DPO (+WVS) and GPT-4o despite substantial human variation; over-diversification on Q64 ($\dr > 1.0$) for unaligned models.}
\label{tab:perq_dr}
\end{table}

\section{Prompt Template}
\label{app:prompt}
All chat models receive the following three-turn chat prompt verbatim. \texttt{\{persona\}}, \texttt{\{question\_text\}}, \texttt{\{min\}}, and \texttt{\{max\}} are filled in per sample; no other prompt engineering, few-shot examples, or chain-of-thought instructions are used. The pre-trained Llama~3.1 70B base has no chat template and receives the same content as a single completion prompt, decoded greedily for at most 8 tokens:

\begin{quote}
\small
\texttt{Survey respondent profile: \{persona\}}\\
\texttt{Question: \{question\_text\}}\\
\texttt{Answer (single integer from \{min\} to \{max\}):}
\end{quote}

\begin{quote}
\small
\texttt{[System]} You are simulating a survey respondent. Answer exactly as instructed, following the specified response format without additional commentary.

\texttt{[User]} You are a survey respondent with the following profile: \{persona\}. Read the question below and answer exactly as this person would. Question: \{question\_text\}. Only return a single integer from \{min\} to \{max\}, nothing else.

\texttt{[Assistant]} \{predicted\_integer\}
\end{quote}

\section{Models and Hyperparameters}
\label{app:models}
Table~\ref{tab:models} lists all evaluated models with training methods, parameter counts, precision, and key hyperparameters. Custom fine-tuning uses LoRA \citep{hu2022lora} (rank 16, $\alpha = 16$) via the Unsloth framework on a single 80\,GB-class NVIDIA GPU (A100 or H100); the 70B model additionally uses 4-bit NF4 quantization \citep{dettmers2023qlora}.

\paragraph{Supervised Fine-Tuning (SFT).} Qwen 3.5 9B trained on 100{,}000 respondent--question pairs from the 15 training questions, drawn as a seeded random sample (seed 3407) of the 331{,}344 training-question pairs and covering all 12 countries (5{,}352--14{,}234 pairs per country); 5\% of the sample is split off for the early-stopping evaluation. Learning rate $2\times 10^{-5}$, 3 epochs, effective batch 8 (per-device 2, gradient accumulation 4), AdamW 8-bit optimizer, BF16 precision, early stopping with patience 3.

\paragraph{DPO --- Qwen 3.5 9B.} Demographic Contrastive DPO following \citet{kolluri2025simulacra}: for each respondent the chosen response is their own answer and the rejected response is a different answer given by another respondent to the same question (a random other value in the rare case that all respondents agree). 100{,}000 such pairs, sampled with a fixed seed from the 15 training questions, learning rate $5\times 10^{-6}$, 1 epoch, $\beta = 0.05$, effective batch 8. The reference model is implicit.

\paragraph{DPO --- Tulu 3 70B.} Replicated on \texttt{allenai/Llama-3.1-Tulu-3-70B-DPO}. 4-bit quantization (NF4, $\sim$35\,GB VRAM), effective batch 8 (per-device 1, gradient accumulation 8). As for Qwen, the reference model is implicit: the trainer scores reference log-probabilities with the LoRA adapter disabled, so no second copy of the 70B weights is loaded.

\paragraph{GRPO.} Qwen 3.5 9B, 8 generations per prompt, $\beta = 0.04$, learning rate $5\times 10^{-6}$, 1 epoch over 50{,}000 training samples, effective batch 16 (per-device 4, gradient accumulation 4). Reward: $+1.0$ for exact match, $-\text{MAE}/\text{scale\_range}$ for near misses, $-1.0$ for parse failures \citep{shao2024deepseekmath}. A second run keeps the same data budget, LoRA configuration, learning rate, $\beta$, and generation count, uses a larger effective batch of 128 (per-device 16, gradient accumulation 8), and replaces the reward with a distribution-shaped one: a sampled answer $a$ earns $p(a \mid \text{cell}) / \max_{a'} p(a' \mid \text{cell})$, its relative frequency among human respondents in the same (country, question) cell, estimated from training questions only.

\begin{table}[h]
\centering
\small
\resizebox{\columnwidth}{!}{
\begin{tabular}{llccl}
\toprule
\textbf{Model} & \textbf{Method} & \textbf{Params} & \textbf{Prec.} & \textbf{Key Config} \\
\midrule
\multicolumn{5}{l}{\textit{Zero-Shot Baselines (no fine-tuning)}} \\
GPT-4o              & Zero-Shot & Prop. & ---    & API, greedy \\
GPT-5.4-mini        & Zero-Shot & Prop. & ---    & API, greedy \\
Claude Opus 4       & Zero-Shot & Prop. & ---    & API, greedy \\
Gemini 2.5 Pro      & Zero-Shot & Prop. & ---    & API (preview endpoint), greedy \\
Gemma 4 31B         & Zero-Shot & 31B   & ---    & API, greedy \\
Qwen 3.5 9B\textsuperscript{$\dagger$} & Zero-Shot & 9B & --- & API, greedy \\
Llama 3.1 70B-Instruct & Zero-Shot & 70B & --- & API, greedy \\
Llama 3.1 70B (base) & Zero-Shot & 70B & --- & Completion prompt, greedy \\
Tulu 3 70B-SFT      & Zero-Shot & 70B   & ---    & AI2 SFT checkpoint \\
Tulu 3 70B-DPO\textsuperscript{$\dagger$} & Zero-Shot & 70B & --- & AI2 DPO checkpoint \\
Tulu 3 70B-RLVR   & Zero-Shot & 70B   & ---    & AI2 final RLVR checkpoint \\
\midrule
\multicolumn{5}{l}{\textit{Custom Fine-Tuned on WVS}} \\
Qwen 3.5 9B\textsuperscript{$\dagger$} & +WVS-SFT  & 9B  & BF16 & lr=$2\!\times\!10^{-5}$, 3 ep. \\
Qwen 3.5 9B\textsuperscript{$\dagger$} & +WVS-DPO  & 9B  & BF16 & lr=$5\!\times\!10^{-6}$, $\beta$=0.05 \\
Tulu 3 70B-DPO\textsuperscript{$\dagger$}        & +WVS-DPO  & 70B & NF4  & lr=$5\!\times\!10^{-6}$, $\beta$=0.05 \\
Qwen 3.5 9B\textsuperscript{$\dagger$} & +WVS-GRPO & 9B  & BF16 & 8 gens/prompt, $\beta$=0.04 \\
Qwen 3.5 9B\textsuperscript{$\dagger$} & +WVS-GRPO (dist.) & 9B & BF16 & cell-frequency reward \\
\bottomrule
\end{tabular}
}
\caption{Models under evaluation. \textsuperscript{$\dagger$}Checkpoints marked with $\dagger$ appear in both sections: evaluated zero-shot (top) and after additional fine-tuning on WVS data (bottom). The proprietary models and the open-weight Qwen, Gemma, and Llama-Instruct checkpoints were queried through OpenRouter, the Llama base and Tulu RLVR checkpoints through the Hugging Face Inference API, and the Tulu SFT and DPO checkpoints and all fine-tuned adapters were run locally with Unsloth; the Qwen $\Delta$ columns in Table~\ref{tab:main} therefore compare an API-served zero-shot run with locally served fine-tuned runs.}
\label{tab:models}
\end{table}
\end{document}